# SDQM: synthetic data quality metric for object detection dataset evaluation

**Ayush Zenith**,[a,*] **Arnold Zumbrun**,[b] **Neel Raut**,[a] **and Jing Lin**[c]
[a]Northeastern University, Khoury College of Computer Sciences, Boston, Massachusetts, United States
[b]Binghamton University, School of Computing, Binghamton, New York, United States
[c]Air Force Research Laboratory, Mission Applications and Infrastructure Section, Rome, New York, United States

**ABSTRACT.** The performance of machine learning models depends heavily on training data. The scarcity of large-scale, well-annotated datasets poses significant challenges in creating robust models. To address this, synthetic data generated through simulations and generative models has emerged as a promising solution, enhancing dataset diversity and improving the performance, reliability, and resilience of models. However, evaluating the quality of this generated data requires an effective metric. We introduce the Synthetic Dataset Quality Metric (SDQM) to assess data quality for object detection tasks without requiring model training to converge. This metric enables more efficient generation and selection of synthetic datasets, addressing a key challenge in resource-constrained object detection tasks. In our experiments, SDQM demonstrated a strong correlation with the mean average precision (mAP) scores of YOLO11, a leading object detection model, whereas previous metrics only exhibited moderate or weak correlations. In addition, it provides actionable insights into improving dataset quality, minimizing the need for costly iterative training. This scalable and efficient metric sets a new standard for evaluating synthetic data. The code for SDQM is available at https://github.com/ayushzenith/SDQM.

## 1 Introduction

There is an increased interest in creating synthetic data to address the data challenges, particularly in domains where collecting and labeling real-world data is expensive and time-consuming. Synthetic data can be generated from various sources, including digital data (e.g., images or videos) produced by physics-based simulated 3D environments or through generative AI. In this paper, we focus on simulation-based synthetic data due to its advantages in physical accuracy, reproducibility, traceability, explainability, and configurability.

However, bridging the gap between synthetic and real data typically involves multiple rounds of testing and refinement. This iterative process is essential to ensure that the synthetic data accurately reflects the characteristics and variability of real-world data, making it useful for training models. Despite the potential benefits, the resources required for effective synthetic data generation, combined with the need for repeated testing and fine-tuning, pose major hurdles to its widespread adoption.

*Address all correspondence to Ayush Zenith, zenith.a@northeastern.edu

We aim to develop an integrated metric—a mathematical formula that combines one or more individual metrics or properties—to assess the usefulness of synthetic datasets for object detection. The key idea is that a higher score on our SDQM metric will correlate with improved model performance when the object detection model is trained on that dataset. This would provide a cost-effective and efficient predictive measure of the dataset's effectiveness, allowing for an estimation of how well a model would perform without the need for exhaustive training and testing.

Although this paper is primarily focused on object detection, this metric could be extended to similar tasks including instance segmentation, object localization, and image classification. Furthermore, as a dataset quality metric, it would serve as a useful benchmark for future research involving the creation of mixed real and synthetic datasets. In summary, our key contributions are as follows:

- We explore various existing data quality metrics for generative AI models and investigate their effect on predicting the performance of object detection models trained on synthetic data.
- We develop a data quality metric, SDQM, for object detection by evaluating the pixel space, spatial space, frequency space, and feature space domain gap using both state-of-the-art AI techniques and various nonparametric distribution comparison techniques, i.e., the Kolmogorov-Smirnov (K-S) test,[1] Anderson-Darling (A-D) test,[2] Kullback-Leibler (KL) divergence,[3] Energy Distance (ED),[4] Wasserstein Distance (WD),[5] Bhattacharyya Distance (BD),[6] and Jensen-Shannon (JS) Divergence.[7]
- We validate the effectiveness of SDQM, demonstrating a strong correlation ($r = 0.87$) between SDQM values and model performance (i.e., mAP50 scores of object detection models trained on synthetic data). The metric's predictive capability reduces reliance on exhaustive training-validation cycles, as shown by consistent performance gains across multiple diverse datasets such as RarePlanes,[8] Dataset of Industrial Metal Objects (DIMO),[9] and WASABI.[10]

## 2 Related Work

Dataset quality has been studied in different contexts, emphasizing data quality metrics and dataset interpretability. In this section, we provide a brief review of the relevant metrics and research on dataset interpretability.

### 2.1 Existing Data Quality Metrics

There exist several approaches to establishing a data metric. Structural Similarity Index Measure (SSIM, MSSIM)[11] was developed to assess the similarity between images by considering changes in structural information. This metric is widely used in image quality assessment. However, this metric is content invariant and thus will not be particularly helpful in comparing the collections of real and synthetic images for the proposed use case. Similarly, Zhang et al.[12] designed Learned Perceptual Image Patch Similarity (LPIPS) to measure the perceptual similarity between images based on deep network features. Although it is useful for comparing the quality of synthetic images with their real counterparts, it is again content-invariant and not helpful in determining the usefulness of images for training an object detection model.

One of the earliest data quality evaluation metrics for GANs is the Inception score (IS).[13] IS assesses the quality of generated images based on the output probabilities of a pre-trained Inception v3 model. Higher scores indicate data are both diverse and recognizable as real objects. However, the metric is biased toward images that resemble those on which the Inception model was trained, and it does not consider the statistics of real-world samples. To improve it, Fréchet Inception Distance (FID)[14] was later introduced to incorporate the statistics of real-world samples to quantify the difference between the feature distributions of real and generated images. Lower FID scores suggest a higher similarity between these distributions. Various FID variants exist, each with its own drawbacks, as discussed in Ref. 15.

Recently, the authors of MAUVE[16] evaluated the similarity between a real and a synthetic data distribution using quantized embeddings and divergence measures. It has shown promise in evaluating the quality of generative data in natural language processing, and by adapting it for image data, it has earned its place as one of the submetrics included in the calculation of SDQM. Generative precision and recall[17] provide a more nuanced evaluation of generative models by measuring the fidelity and diversity of generated samples, although MAUVE seems to outperform these metrics when it comes to comparing two distributions in the generative evaluation space. Alaa et al.[18] improved the generative precision and recall metrics by designing the $\alpha$-precision, $\beta$-recall, and authenticity metrics to measure fidelity, diversity, and authenticity, respectively.

- $\alpha$-Precision:[18] Determines the extent to which synthetic data aligns with the dense regions of the real data distribution. High $\alpha$-precision ensures the dataset includes realistic and plausible samples without focusing excessively on outliers or edge cases.
- $\beta$-Recall:[18] Measures the diversity of synthetic data by evaluating the fraction of real data distribution covered by the generative model. High $\beta$-recall implies that the synthetic dataset is capturing a broad range of modes from the real data, avoiding mode collapse or narrow sampling.
- Authenticity:[18] Assesses the novelty of synthetic samples by determining how many of them are unique and not near duplicates or direct copies of real training data. High authenticity indicates the synthetic dataset is capable of generating genuinely new and diverse samples.

These individual metrics, although valuable, cannot address the complexities of evaluating data and are incapable of metricizing quality, which we define as a direct correlation of model performance.

### 2.2 Dataset Interpretability

The previously mentioned metrics focus primarily on evaluating data quality through data distributions. Dataset interpretability, however, involves assessing the value of individual data points or subsets within a larger dataset, identifying which specific points or subsets most contribute to improving model performance. The concept of $\mathcal{V}$-usable information[19] quantifies the useful information a dataset provides for a specific task, offering a way to assess dataset difficulty (how challenging it is for a model to learn from it). We draw upon the framework introduced by Ethayarajh et al.[19] to incorporate dataset interpretability into our metric. Data Maps[20] is another proposed technique to analyze training dynamics and categorize data points based on their difficulty and consistency. This method provides valuable insights into dataset quality and helps identify problematic examples that may hinder model training.

Together, these related works lay a solid foundation for developing a comprehensive dataset quality metric tailored to synthetic data in object detection tasks. They also establish benchmarks against which our proposed metrics can be compared. By extending and integrating these existing methods, we aim to create a robust metric that not only quantifies dataset quality but also guides the generation of high-quality synthetic datasets.

## 3 Methodology

In this section, we present our methodology for developing SDQM, illustrated in Fig. 1. SDQM is an integration of numerous independent components (or submetrics). Many of these submetrics have been previously defined in the literature and are detailed in Sec. 2. However, we also introduce new submetrics and extend existing ones for object detection in Sec. 3.1. These novel submetrics are designed to assess the utility of images for training object detection models, complementing the existing metrics that primarily evaluate data quality or alignment with distribution. After introducing the novel submetrics, we detail the process of integrating all submetrics together and creating SDQM in Sec. 3.3.

### 3.1 Individual Metrics for Dataset Quality Evaluation

We begin by reviewing existing data quality metrics and assessing their impact on predicting the performance of object detection models trained on synthetic data. We then select a range of established metrics and propose new methods for calculating additional metrics that complement the existing ones. Figure 2 shows the moderate correlation with mAP50 (first row) of the components. Notably, the strong correlation between MAUVE and FI (stemming from the fact that they measure similar characteristics) indicates that we may not need both of them. Furthermore, Fig. 2 also illustrates the correlation (all $\geq 0.62$) among $\alpha$-precision, $\beta$-recall, and authenticity, demonstrating the interconnection between the metrics. In addition to presenting raw correlations, the figure groups metrics according to their correlation clusters (blue, orange, and green row and column colors). This clustering highlights which metrics tend to behave similarly and therefore may offer redundant or complementary information. We will later analyze these metrics to explore better integration methods for higher correlation with mAP50.

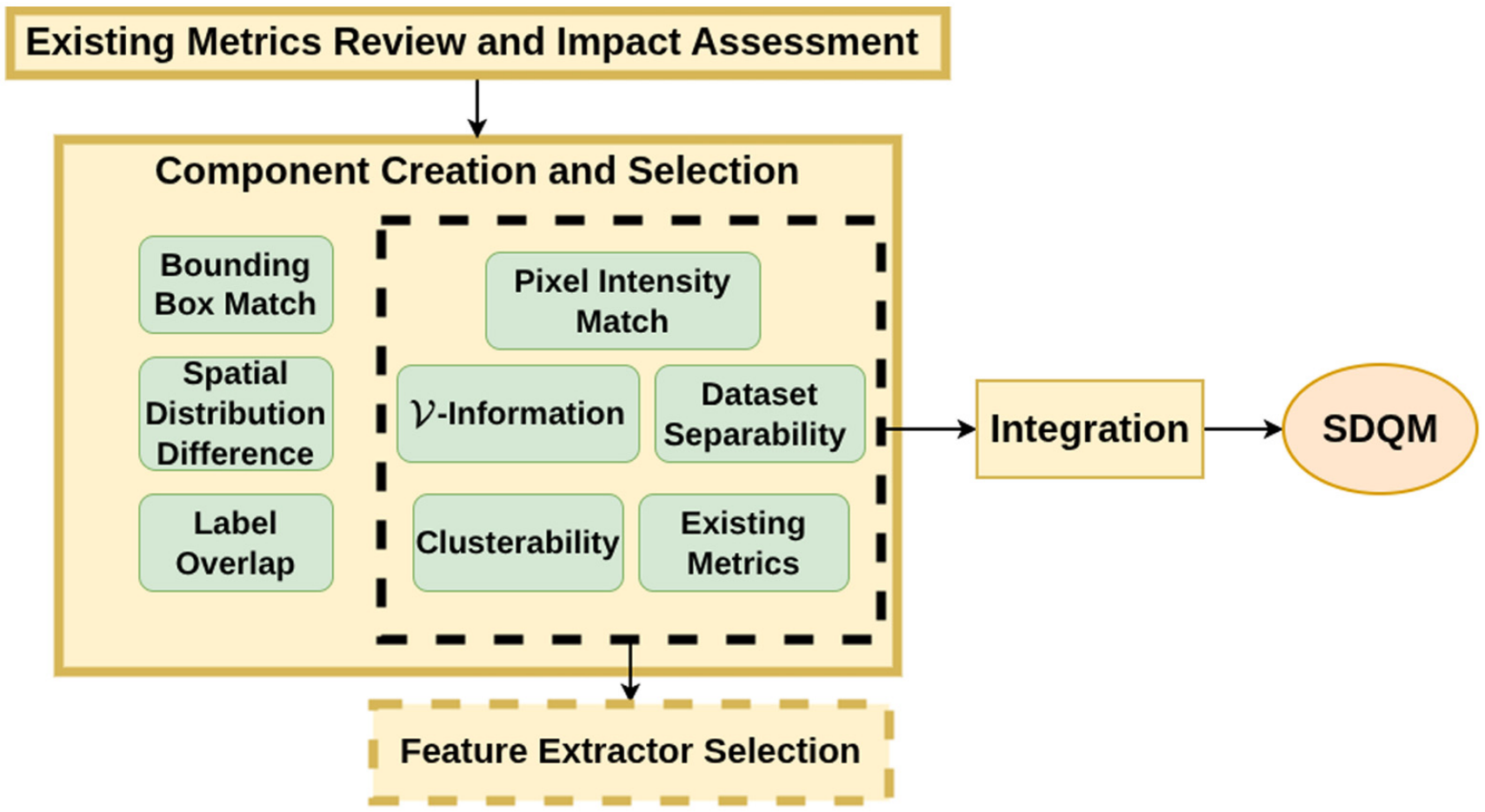

**Fig. 1** Overview. The process begins with a review and assessment of existing metrics. This is followed by a detailed component creation and selection phase, which evaluates various factors, including Bounding Box Match, Spatial Distribution Difference, Pixel Intensity Match, and Clusterability, among others. The selected components are then integrated to form the final SDQM, along with a separate process for selecting the feature extractor.

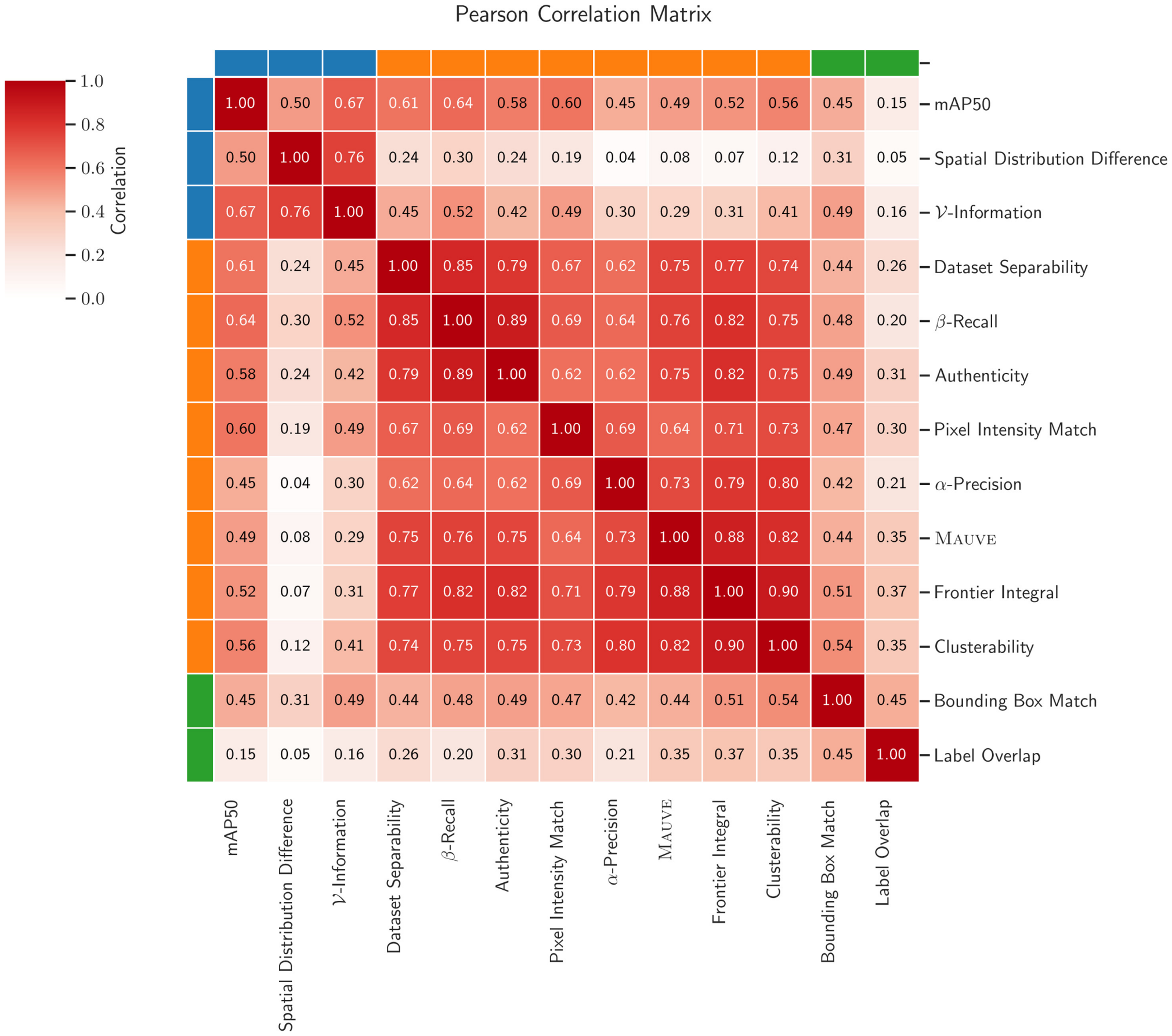

**Fig. 2** Pearson correlation matrix of evaluated metrics against downstream mAP50. The matrix axes are ordered via hierarchical clustering to juxtapose metrics with high statistical similarity. The colored sidebars indicate the metric clusters.

Although these metrics are suitable for our purpose, each has limitations. We propose two additional metrics—dataset separability and clusterability—to better assess synthetic data embedding space overlap. The dataset separability component utilizes AutoKeras[21] to choose a neural network architecture and train the network to classify embeddings as real or synthetic. To measure the level of ease with which the model separated the real and synthetic embeddings, we record the evaluation accuracy of the highest-performing model. Theoretically, a sufficiently complex model can separate any set of real and synthetic embeddings with a high evaluation accuracy; therefore, a model's evaluation accuracy must be juxtaposed with its complexity for an accurate measure of dataset separability. To track the balance between effectiveness and efficiency, we record the parameter count of the model selected by AutoKeras in addition to its evaluation accuracy. These two measures are weighted appropriately along with other metrics in Sec. 3.3.

To assess clusterability, we employ the $k$-means clustering algorithm to jointly cluster the real $R$ and synthetic $S$ embeddings. The log cluster metric equation,

$$\mathrm{LC}(R, S) = \log\left(\frac{1}{k}\sum_{i=1}^{k}\left[\frac{n_i^R}{n_i} - \frac{n^R}{n}\right]^2\right), \tag{1}$$

accentuates smaller differences between the data points, where $k$ is the number of clusters that separate the combined synthetic $S$ and real $R$ embeddings, $n$ is the dataset size (i.e., $n = |S| + |R|$), $n^R$ is the total number of real images (i.e., $n^R = |R|$), $n_i$ is the total number of data points assigned to cluster $i$, and $n_i^R$ is the number of real data points assigned to cluster $i$. In simpler terms, we compute the average of the squared differences between the actual proportion of real data points in each cluster and the expected proportion of real images across all clusters. If the clusters are inseparable, the expected proportion is $\frac{n^R}{n}$. Figure 2 shows that dataset separability and clusterability each correlate moderately with mAP50; furthermore, the two components themselves correlate strongly with each other and certain other components. Section 3.3 further investigates this.

To evaluate the objects and characteristics in the synthetic dataset, we quantify the Label Overlap between synthetic and real imagery. Using the annotations and metadata of real and synthetic datasets, we can create probability distributions for each and use probability distribution comparison metrics. In each case, we construct a probability distribution reflecting the number and categories of objects within an image. In addition, when the dataset includes informative metadata—such as weather conditions, biome, or physical location—we incorporate this metadata into the distribution used for comparison.

Spatial distribution difference compares the distributions of objects in the real imagery and its synthetic counterpart. The more similar the spatial distribution of synthetic data is to real data, the lower the spatial distribution score will be. We calculate the spatial distribution score by initializing a spatial distribution heatmap as a matrix of size $w \times h$, where $w$ is the image width and $h$ is the image height. Each position in the matrix is incremented by one for each object that overlaps with the pixel corresponding to that position. The real heatmap is then compared with the synthetic heatmap using root mean squared error (RMSE). Before the RMSE operation, a mean pooling operation reduces the size of the heatmap by a factor of eight; the pooling forces RMSE to work on a more global scale and adds smaller penalties for more localized differences.

In addition, we evaluate bounding box geometry—specifically aspect ratios, diagonal lengths, and areas—using nonparametric tests to derive a comprehensive Bounding Box Match score. We then analyze the pixel intensity in all images. Comparing the overall pixel intensity distributions of the synthetic and real datasets, we derive our Pixel Intensity Match component. To ensure comparable pixel intensities across RGB channels in synthetic versus real data, we calculate per-image channel distributions and apply several nonparametric tests.

Finally, we calculate and metricize the interpretability of the synthetic dataset by accounting for the object detection model family $\mathcal{V}$. We use the foundation laid in $\mathcal{V}$-usable information[19] and extend it for object detection models. We call our extension simply $\mathcal{V}$-information, which measures the difficulty of the dataset by analyzing how easily a model can learn from the synthetic data and apply it to the target domain. This will help us understand how to close the domain gap. The $\mathcal{V}$-information equation is

$$I_V(X \rightarrow Y) = H_V(Y) - H_V(Y|X). \tag{2}$$

To calculate the conditional and predictive entropy $H_V$ for the above-described variants in an efficient manner on the YOLO11[22] family, the following implementation is employed:

- Conditional entropy is defined as

$$H_{\mathcal{V}}(Y|X) = \inf_{f \in \mathcal{V}} \mathbb{E}[-\log_2 f[X](Y)], \tag{3}$$

  where $f[X](y)$ indicates that the object detection model $f \in \mathcal{V}$ is conditioned on the training set $X$ (i.e., trained on $X$), whereas the argument $y$ represents the input data for inference. To compute the conditional entropy, we fine-tuned a model, in our case, the pretrained YOLO11n checkpoint, on the synthetic dataset, while freezing the backbone layers for $E$ epochs, ensuring the number of epochs was greater than the warm-up phase yet still kept relatively low. We opted for YOLO11n as it strikes an optimal balance between minimizing computation time and allowing us to effectively assess the model's performance and knowledge before and after training. In our experiments, YOLO11n was trained for $E = 10$ epochs before validation on the real dataset, during which we calculated the conditional entropy.
- Predictive entropy is defined as

$$H_{\mathcal{V}}(Y) = \inf_{f \in \mathcal{V}} \mathbb{E}[-\log_2 f[\varnothing](Y)]. \tag{4}$$

  To calculate the predictive entropy, we use the pretrained YOLO11n model, which has been trained on the COCO dataset,[23] and validate its performance on a real dataset. During validation, we map the model's predicted class labels to those in the real dataset (for instance, in RarePlanes, class 0 is "airplane," whereas in COCO, class 5 is airplane). Predictive entropy is then computed using these mapped matches.

As shown in Fig. 2, our extended $\mathcal{V}$-information exhibits the strongest individual correlation with mAP50 compared with all others discussed. We will further improve it to obtain the SDQM metric. Notably, the MAUVE, FI score, clusterability, $\mathcal{V}$-information, $\alpha$-precision, $\beta$-recall, and authenticity metrics all require embeddings extracted from images to work effectively. Next, we outline the process of selecting the feature extraction model.

### 3.2 Evaluating Feature Extraction Models

Feature extraction forms the backbone of many metrics. To obtain optimal results, selecting a high-quality model for feature extraction is crucial. Below are the feature extractors we have chosen for further exploration and evaluation.

1. GroundingDINO-tiny:[24] a compact open-set detector, ideal for efficiently aligning features between real and synthetic data. The model was modified to use only the encoder, which takes in an image and a text prompt and outputs features in 256 dimensions.
2. CLIP-ViT-B/32:[25] A multimodal vision transformer excelling at zero-shot generalization by encoding inputs into a shared space. The ViT-B/32 variant processes $32 \times 32$ patches into 512-dimensional embeddings.
3. DinoV2-small:[26] A robust vision transformer trained using self-supervised learning methods. Known for its strong performance, DinoV2-small excels at capturing and extracting semantic features across diverse datasets, all without the need for labeled data.

Our selection of the initial candidate pool was driven by

- Architectural diversity: Included distinct models to compare different approaches (e.g., CLIP for global semantics versus GroundingDino for object-level semantics).

- Computational efficiency: Prioritized "tiny" or "small" variants to ensure the metric remains fast and lightweight.
- Availability: Selected architectures with readily available pretrained weights.

We evaluate the feature extractor using the DIMO dataset, which provides real and synthetic image pairs. The dataset features objects with varying shapes and angles in a controlled environment, making it ideal for assessing feature extraction performance. For each real-synthetic pair, we compute cosine similarity and Euclidean distance between the feature representations produced by different models. Cosine similarity measures angular alignment (values near 1 indicate similarity), whereas Euclidean distance captures absolute spatial separation (lower values indicate proximity). Together, these metrics offer complementary insights into how effectively a model maintains both semantic and spatial consistency between synthetic and real data. The models are ranked based on their average and cumulative performance across all pairs. The model with the highest average cosine similarity, lowest average Euclidean distance, and best overall cumulative scores is selected as the most effective feature extractor for preserving fidelity between real and synthetic data.

Table 1 presents the evaluation results for three models: GroundingDino-tiny, CLIP-ViT-B/32, and DinoV2-small. The models differ significantly in their ability to maintain consistency between synthetic and real data, as evidenced by variations in cosine similarity and Euclidean distance statistics. GroundingDino-tiny yielded the most robust feature representations among the three variants tested. This evaluation demonstrates the model's ability to consistently capture meaningful features across the diverse conditions present in the DIMO dataset.

**Table 1** Feature extraction model evaluation.

| Model | Cos Sim Sum | Mean Cos Sim | Euc. Dist. Sum | Mean Euc. Dist. |
|---|---|---|---|---|
| GroundingDino-tiny | 7707.237835 | 0.989757 | 760.091262 | 0.097610 |
| CLIP-ViT-B/32 | 6949.146740 | 0.892404 | 40393.930023 | 5.187355 |
| DinoV2-small | 6566.421186 | 0.843254 | 129133.357791 | 16.583197 |

**Table 2** Description of all parameters, functions, and variables in Algorithm 1.

| Parameter | Description |
|---|---|
| $\mathcal{X}$ | Images from real dataset |
| $\mathcal{X}'$ | Images from synthetic dataset |
| $k_l, k_u$ | Lower and upper bounds for subset size |
| $g$ | Number of generations |
| $n$ | Number of values to optimize for |
| $\mathcal{M}$ | Set of metrics to use |
| $m$ | a metric in $\mathcal{M}$ |
| $p$ | Population size |
| $P_m$ | Mutation probability |
| $P_c$ | Crossover probability |
| $\mathrm{dist}(a, b, c)$ | Distance of $c$ outside $[a, b]$. (If $c$ is within $[a, b]$, then $\mathrm{dist}(a, b, c) = 0$.) $\mathrm{dist}(a,b,c) = \begin{cases} a - c, & \text{if } c < a \\ c - b, & \text{if } c > b \\ 0, & \text{otherwise} \end{cases}$ |
| $\varphi$ | Embedding model |

### 3.3 SDQM: Integrated Metric Development

We employ an evolutionary algorithm[27] to identify and remove redundant subcomponents. The algorithm selects subset pairs exhibiting a range of similarity levels to undergo correlation analysis. For each submetric $m$ outlined in Sec. 3.1 and for each of the $n$ values equally spaced across the range of $m$, i.e., $L = \{m_j : m_{j+2} - m_{j+1} = m_{j+1} - m_j\}_{j=1}^{n}$, we execute the algorithm. The fitness function computes the deviation between the individual's submetric value and the target optimization value. Furthermore, as described in Algorithm 1, a penalty is imposed proportional to the deviation of subset sizes from the target range. Table 2 summarizes the parameters and variables utilized in Algorithm 1.

Following an exploratory analysis of subcomponent interactions and their correlation with mAP50, we identified Random Forest regression as the optimal modeling approach (detailed in the appendix). This method effectively handles multivariate dependencies and identifies the most influential components, which we discuss further in this section.

The Bounding Box Match, Pixel Intensity Match, and Label Overlap components each comprise a set of nonparametric statistical tests; therefore, we analyze the contributions of each of

**Algorithm 1** Dataset selection.

**Input:** $\mathcal{X}$, $\mathcal{X}'$, $k_l$, $k_u$, $g$, $n$, $P_m$, and $P_c$
$\Phi = \varphi(\mathcal{X}) \cup \varphi(\mathcal{X}')$
**for** $m \in \mathcal{M}$ **do**
  Select $n$ values equally spaced across the range of $m$, i.e., $L = \{m_j : m_{j+2} - m_{j+1} = m_{j+1} - m_j\}_{j=1}^{n}$
  **for** $m_j \in L$ **do**
    $\mathcal{P} = \varnothing$.
    **for** $i = 1$ to $p$ **do**
      Randomly select $s_1^i, s_2^i \in [k_l, k_u]$.
      Randomly generate $D^i = (D_1^i, D_2^i)$ such that $D_1^i, D_2^i \subset \Phi$, $D_1^i \cap D_2^i = \varnothing$, $|D_1^i| = s_1^i$, and $|D_2^i| = s_2^i$
      $\mathcal{P} = \mathcal{P} \cup D^i$
      $f(D^i, m, m_j) = |m_j - m(D^i)| + \frac{\text{dist}(k_l, k_u, |D_1^i|) + \text{dist}(k_l, k_u, |D_2^i|)}{\max(k_u - k_l, 1)} \cdot \text{Range}(m)$
    **end for**
    **for** each generation **do**
      **if** the minimum individual fitness in the population is less than 0.005 **then**
        Select the individual with this fitness.
        Continue to the $m_{j+1}$.
      **end if**
      Create offspring $D$ from the population using crossover with probability $P_c$ and mutation with probability $P_m$
      Evaluate $f(D, m, m_j)$
    **end for**
    **if** no individual has been selected **then**
      Select the individual with the minimum fitness
    **end if**
  **end for**
**end for**

these tests to the performance of each submetric and select the best performing statistical test for each submetric, i.e., A-D Statistic for Pixel Intensity Match, Energy Distance for Bounding Box Match, and K-S Statistic for Label Overlap.

For each of the other components, we conduct multivariate regression using linear, quadratic, and interaction terms to examine their interrelationships. The selection of terms is based on p-values from the regression results. For example, in the Dataset Separability component, which has the number of parameters ($p$) and validation accuracy ($a$) as subcomponents, the resulting regression analysis assigns a negative weight to $a$ and a positive weight to $p$. This reflects the notion that synthetic data becomes easier to distinguish from real data as $a$ increases with the same $p$, but harder to separate when $p$ is lower with the same $a$. This finding links higher Dataset Separability to higher mAP50 scores, confirming that synthetic data closely resembling real data leads to better performance.

To illustrate how the submetrics relate to each other and the performance of the empirical model, the correlation matrix in Fig. 2 was created using data points produced by Algorithm 1. Observing that MAUVE correlates highly with Frontier Integral, authenticity correlates highly with $\beta$-recall, and clusterability correlates highly with FI and MAUVE, we ran the backward feature reduction method[28] on the data and confirmed that the removal of MAUVE, authenticity, and clusterability does not reduce linear regression performance and only slightly reduces random forest performance. We remove these three submetrics from the data before proceeding with the regression analysis in Sec. 4.

Figure 2 indicates that the submetrics with the highest correlation with the empirical model performance are $\mathcal{V}$-information, $\beta$-recall, Dataset Separability, and Pixel Intensity Match. Label Overlap correlated the least with model performance. The correlations between each of our selected submetrics and mAP50 provide valuable insights into their effectiveness and reveal certain intimate characteristics of object detection models. For example, the weak correlation between Label Overlap and mAP50 indicates that YOLO11 can effectively learn to disregard variations in object counts and certain other annotation attributes within datasets. Conversely, the high correlation of the Pixel Intensity Match submetric emphasizes its importance in synthetic-real domain gap analysis. Finally, our $\mathcal{V}$-information submetric shows the strongest correlation with model performance, underscoring its value in evaluating training data. By integrating these submetrics through a random forest, we generate the proposed metric. Detailed information is provided in Sec. 4.

## 4 Experiments

This section presents extensive experiments to validate the proposed metric, investigating its correlation with the model performance. In this study, we use mAP50 as a proxy for "ground-truth" quality, following the Train on Synthetic, Test on Real (TSTR) evaluation protocol,[29] the established method for assessing synthetic datasets in computer vision. Although we recognize that mAP50 is tied to how well the real-world test set reflects deployment conditions, we argue that because the main objective of synthetic data is to improve downstream detection performance, aligning with the standard detection benchmark (mAP50) is the most practical approach. To further address this limitation, we use Algorithm 1 to select multiple train-test dataset pairs. Hence, the result is not specific to a single test set. Accordingly, we use mAP50 not as a proxy for visual realism, but as the authoritative measure of task utility—i.e., how effectively synthetic data supports object-detection training—and thus as the primary indicator of detection performance in this work.

### 4.1 Datasets

To ensure robust verification, we collect a large number of data points by employing an evolutionary algorithm to select subsets from three synthetic-real dataset pairs. This robust methodology enables a thorough evaluation of the metric's effectiveness across various challenging scenarios, enhancing its validity and practical applicability.

- Rareplanes[8]—a large-scale collection of real and synthetic overhead imagery specifically designed to evaluate the effectiveness of synthetic data in training object detection models.

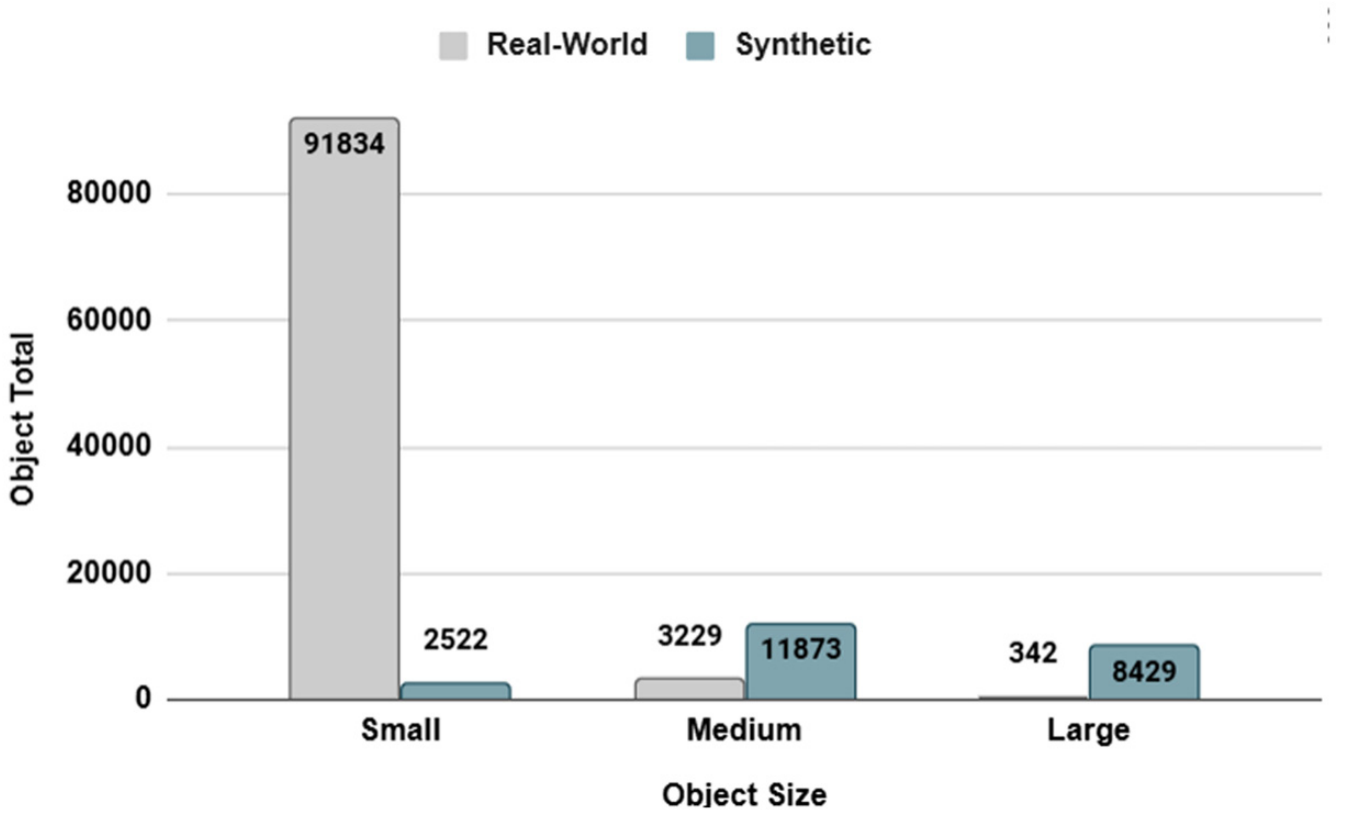

**Fig. 3** WASABI object size and count. Note: The definitions for "small," "medium," and "large" are adopted from Ref. 23.

Its high-quality annotations and diverse features make it a valuable tool for researching how synthetic data affects model performance. The dataset includes

- 130 different real locations and fifteen synthetic locations
- Five distinct synthetic biomes: alpine, arctic, temperate evergreen forests, grasslands, and tundra
- Weather information for real and synthetic data
- Other information such as sunlight intensity, time of day, camera angles, and real-world dates and times

- Dataset of Industrial Metal Objects (DIMO)[9]—a high-resolution collection designed to advance research in object detection, including evaluating the impact of synthetic variations on model generalization. The dataset consists of
  - Real-world data: 31,200 images from 600 scenes
  - Synthetic data: Generated using Unity, the dataset includes 553,800 images across 42,600 scenes. The synthetic data in the DIMO dataset replicates real-world lighting, object poses, and textures. Textures are designed to simulate real-world effects, such as scratches, using texture synthesis algorithms.
- WASABI[10]—an overhead collection featuring various vehicle and personnel labels. This research centers exclusively on vehicle detection. The dataset comprises about 100,000 real images from 519 videos, with object count distribution illustrated in Fig. 3. It also features a synthetic collection of over 20,000 images, preserving the resolution of the original images.

### 4.2 Dataset Partitioning

We first divide the data into distinct training, evaluation, and test splits to prevent scene leakage, then apply Algorithm 1 to each. As the algorithm only converged reliably for $\alpha$-precision, $\beta$-recall, and authenticity, we restrict selection to these three submetrics. By combining three embedding models, three submetrics, and 11 equally spaced thresholds, we generate a diverse collection of 364 training and 64 evaluation subsets.

### 4.3 Model Training

A YOLO11n model was trained for ten epochs on each subset pair and evaluated on real test data. The evolutionarily selected subsets yielded training mAP50 values of 0.31 to 0.79, compared with 0.06 to 0.40 on evaluation data; this sharp decline underscores the domain gap between synthetic and real data. To introduce size diversity and expand the performance spectrum, we incorporated randomly paired subsets. This addition broadened the evaluation mAP50 distribution to a range of 0.06 to 0.79.

### 4.4 SDQM Development

After the above procedures, a regression can be performed with the submetric values between a subset pair as inputs and the mAP50 score of the respective YOLO11n model as the output. These values for each dataset pair are collected and shuffled before splitting into train and evaluation sets. To evaluate regression robustness, we performed 10-fold cross-validation. A sample result from a random train split and a random evaluation split is presented in Fig. 4; the trial yielded a Pearson correlation coefficient of 0.87 and a Spearman correlation coefficient of 0.68 on the evaluation split. Furthermore, the regression analysis identified Pixel Intensity Analysis, $\beta$-recall, and $\mathcal{V}$-information as the most significant predictors.

As SDQM is a novel approach, we compare its performance against the closest existing standards: generative model evaluation methods. Table 3 presents the Pearson correlation values of existing metrics and SDQM when plotted against the mAP50 values obtained in our subset selection and YOLO11 model training procedure. Unlike existing methods, which reach only moderate correlation levels, SDQM demonstrates a strong correlation with mAP50. Notably, authenticity has a negative correlation with mAP50, as authenticity is a metric originally designed for generative modeling to assess sample uniqueness and ensure that synthetic data are not merely memorized copies of the training set. However, in the context of training object detectors, "uniqueness" does not directly translate to "utility." A synthetic dataset can be highly "authentic" (unique) but fail to capture the necessary features of the real domain (domain gap), resulting in poor downstream performance. Furthermore, our dataset selection process (Algorithm 1) inherently manages diversity constraints, rendering the authenticity metric redundant for this specific purpose. The observed negative correlation is negligible in magnitude, confirming that authenticity is a weak and unreliable predictor of synthetic data training utility for this task.

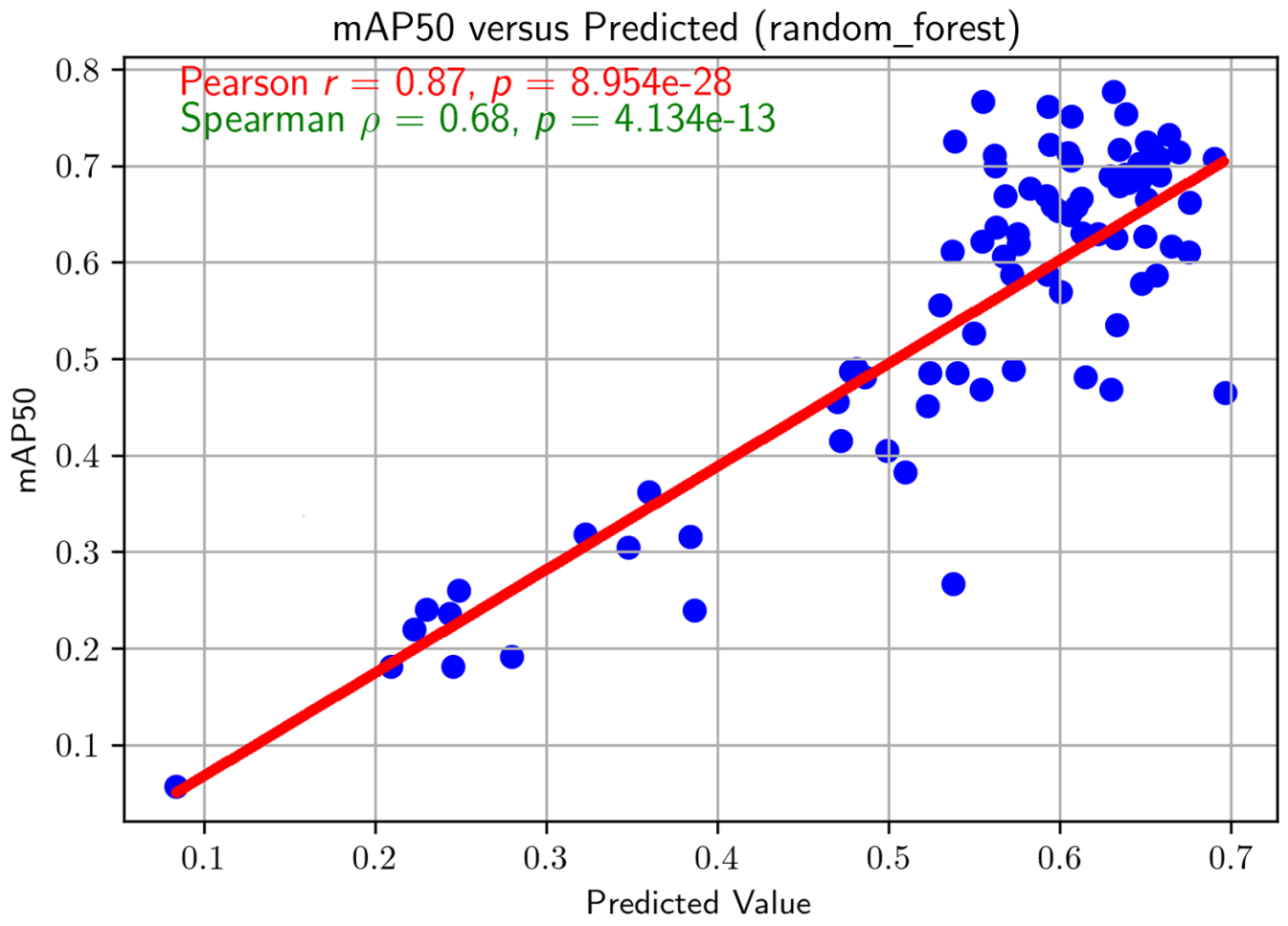

**Fig. 4** Validation datapoints with SDQM values calculated from **Random Forest** coefficients versus YOLO11n mAP50 scores.

**Table 3** Correlation of alternative metrics and SDQM with mAP50 scores.

| | mAP50 |
|---|---|
| Metric | Pearson correlation |
| $\alpha$-Precision[18] | 0.4477 |
| $\beta$-Recall[18] | 0.5865 |
| Authenticity[18] | −0.3482 |
| MAUVE[16] | 0.5651 |
| **SDQM (ours)** | **0.8719** |

Finally, to evaluate a synthetic dataset using mAP50, one must perform the full "Train on Synthetic, Test on Real" protocol. Although both are dependent on dataset size, SDQM consistently offers a substantial speed advantage over the training-intensive mAP50 metric. For example, in our benchmark setup (NVIDIA T4, DIMO Small dataset), calculating the SDQM score takes just 6 min and 21 s, whereas training a YOLO11n model (500 epochs) to obtain a comparable mAP50 score exceeds 11 h. This represents an efficiency improvement of approximately two orders of magnitude.

## 5 Conclusion

In conclusion, SDQM provides a simple approach to evaluating synthetic dataset quality for object detection tasks. By combining multiple submetrics and subcomponents that address critical aspects of dataset quality—including coverage, diversity, domain similarity, and interpretability—SDQM predicts the efficacy of synthetic datasets in improving model performance in a fraction of the compute time of full model training. By aggregating coverage, diversity, and domain similarity, SDQM achieves a Pearson correlation of 0.87 with downstream YOLO11 performance, significantly surpassing baselines such as MAUVE. This strong correlation validates SDQM as a reliable tool for efficient dataset selection, offering deep insights into data quality while enabling scalable synthetic data generation for object detection tasks.

## 6 Appendix A: Existing Submetric Evaluation: $\alpha$-Precision, $\beta$-Recall, and Authenticity

The results presented in Table 4 highlight the $\alpha$-precision, $\beta$-recall, and authenticity scores for different datasets across three models: DinoV2, Grounding-Dino, and ViT. These metrics are crucial for understanding how well synthetic and real datasets perform in terms of precision, recall, and authenticity when paired with different models.

For our datasets, $\alpha$-precision demonstrated the greatest dispersion among the three metrics. DIMO and its synthetic counterpart had the highest $\alpha$-precision scores, suggesting the synthetic counterpart matched the real data distribution well. $\beta$-recall, which is low across all tests, suggests that a significant portion of the real data fell outside the synthetic distributions in our dataset. Finally, as expected, because our synthetic datasets are all physics-based, authenticity indicates that synthetic datapoints were not "stolen" from real datapoints.

**Table 4** $\alpha$-precision, $\beta$-recall, and authenticity scores of the various synthetic and real-world dataset pairs.

| Dataset | Model | $\alpha$-Precision | $\beta$-Recall | Authenticity |
|---|---|---|---|---|
| DIMO | DinoV2-small | 0.394860 | 0.051795 | 0.996389 |
| DIMO | Grounding-Dino | 0.827369 | 0.182025 | 0.983838 |
| DIMO | ViT | 0.682937 | 0.001035 | 0.997654 |
| DIMO-Counterpart | DinoV2-small | 0.602352 | 0.029452 | 0.947220 |
| DIMO-Counterpart | Grounding-Dino | 0.938879 | 0.078241 | 0.902401 |
| DIMO-Counterpart | ViT | 0.776746 | 0 | 0.979068 |
| WASABI | DinoV2-small | 0.149050 | 0 | 1 |
| WASABI | Grounding-Dino | 0.087010 | 0 | 1 |
| WASABI | ViT | 0.041659 | 0 | 1 |
| RarePlanes | DinoV2-small | 0.416685 | 0 | 0.997577 |
| RarePlanes | Grounding-Dino | 0.683583 | 0.185365 | 0.921733 |
| RarePlanes | ViT | 0.466974 | 0 | 0.985980 |

## 7 Appendix B: Existing Submetric Evaluation: MAUVE, FI, MAUVE*, and FI*

Table 5 shows MAUVE and FI scores as well as the smoothed scores (indicated by *) for different synthetic and real dataset pairs using different feature extractors. As illustrated, the metric scores are sensitive to the choice of feature extractor. Feature extractors such as DinoV2-small, CLIP-ViT-B/32, and GroundingDino-tiny produce varying scores for the same dataset pairs. This variability underscores the metrics' dependence on the distinct feature representations generated by each model. This highlights the need to tailor feature extractors to the data and to standardize evaluation pipelines for reliable, consistent comparisons.

**Table 5** MAUVE and Frontier Integral scores (along with smoothed variants indicated by *) of the various synthetic and real-world dataset pairs.

| Model | Dataset *P* | Dataset *Q* | MAUVE | MAUVE* | FI | FI* |
|---|---|---|---|---|---|---|
| DinoV2-small | RP: R Tr | RP: R Val | 0.610339 | 0.662745 | 0.158725 | 0.140987 |
| DinoV2-small | RP: R Tr | RP: Synth | 0.010485 | 0.020507 | 0.846375 | 0.734662 |
| DinoV2-small | W: R Tr | W: R Val | 0.008292 | 0.017457 | 0.886670 | 0.761341 |
| DinoV2-small | W: R Tr | W: Synth | 0.004073 | 0.009237 | 0.999945 | 0.862455 |
| DinoV2-small | DIMO: R | DIMO: Synth | 0.057809 | 0.093659 | 0.573631 | 0.491431 |
| DinoV2-small | DIMO: R | D | 0.011015 | 0.025800 | 0.841786 | 0.697453 |
| CLIP-ViT-B/32 | RP: R Tr | RP: R Val | 0.677107 | 0.729791 | 0.136138 | 0.119079 |
| CLIP-ViT-B/32 | RP: R Tr | RP: Synth | 0.011446 | 0.021393 | 0.833652 | 0.729436 |
| CLIP-ViT-B/32 | W: R Tr | W: R Val | 0.008366 | 0.017573 | 0.885283 | 0.760340 |
| CLIP-ViT-B/32 | W: R Tr | W: Synth | 0.004075 | 0.009248 | 0.999894 | 0.862232 |
| CLIP-ViT-B/32 | DIMO: R | DIMO: Synth | 0.052533 | 0.083358 | 0.589168 | 0.510899 |
| CLIP-ViT-B/32 | DIMO: R | D | 0.005649 | 0.014801 | 0.947456 | 0.784795 |
| GD text: "plane. aeroplane" | RP: R Tr | RP: R Val | 0.898451 | 0.913681 | 0.062346 | 0.056551 |
| GD text: "plane. aeroplane" | RP: R Tr | RP: Synth | 0.085130 | 0.125567 | 0.507735 | 0.441276 |
| GD text: "plane. aeroplane. jet. airplane. flight" | RP: R Tr | RP: R Val | 0.882684 | 0.899710 | 0.068129 | 0.061894 |
| GD text: "plane. aeroplane. jet. airplane. flight" | RP: R Tr | RP: Synth | 0.065192 | 0.097946 | 0.550988 | 0.481972 |
| GD text: "vehicle" | W: R Tr | W: R Val | 0.009586 | 0.019745 | 0.863149 | 0.741495 |
| GD text: "vehicle" | W: R Tr | W: Synth | 0.006827 | 0.013952 | 0.917200 | 0.797097 |
| GD text: "vehicle. car. jeep. truck" | W: R Tr | W: R Val | 0.010600 | 0.022013 | 0.846370 | 0.723660 |
| GD text: "vehicle. car. jeep. truck" | W: R Tr | W: Synth | 0.004072 | 0.009397 | 0.999999 | 0.859747 |
| GD text: "metal cube. rectangular tray. circular metal piece. cylindrical metal disc. metal rod. metal pipe" | DIMO: R | DIMO: Synth | 0.212445 | 0.285720 | 0.354814 | 0.301897 |
| GD text: "metal cube. rectangular tray. circular metal piece. cylindrical metal disc. metal rod. metal pipe" | DIMO: R | D | 0.073652 | 0.120384 | 0.533366 | 0.448752 |

RP, RarePlanes; GD, GroundingDino-tiny; W, WASABI; R, Real; Synth, Synthetic; D, DIMO Synth Counterpart; Tr, Train; and Val, Validation

## 8 Appendix C: Comparison of Regression Methods

Because the random forest regression method yielded the best correlation with our data, we chose to display it in the main paper section. In addition, we present the results of five other regression methods: linear regression (Fig. 5), ridge regression (Fig. 6), XGBoost (Fig. 7), decision tree (Fig. 8), and lasso (Fig. 9), respectively. Interestingly, all of these methods, with the exception of lasso regression, reach Pearson correlation coefficients within 0.05 of random forest.

Figure 5 displays predictions using linear regression. Attesting to the robustness of SDQM's underlying submetrics, this simple linear regression model still performed well on our data. Consequently, linear regression serves as a viable alternative for calculating SDQM when a simpler, more interpretable model is preferred.

Figure 6 displays ridge regression results. The performance of ridge regression, linear regression with an $L2$ regularization term, achieves the same performance as standard linear regression. It indicates that the $L2$ regularization applied in Ridge regression had a negligible effect on the fitted curve. This confirms that the proposed metric does not suffer from high variance or overfitting, as the performance remains consistent even when regularization is introduced.

Figure 7 displays the performance of XGBoost. This regression method achieved the same Spearman correlation coefficient as random forest and a Pearson correlation coefficient only 0.01 less than random forest; however, we chose random forest due to its simplicity and robustness.

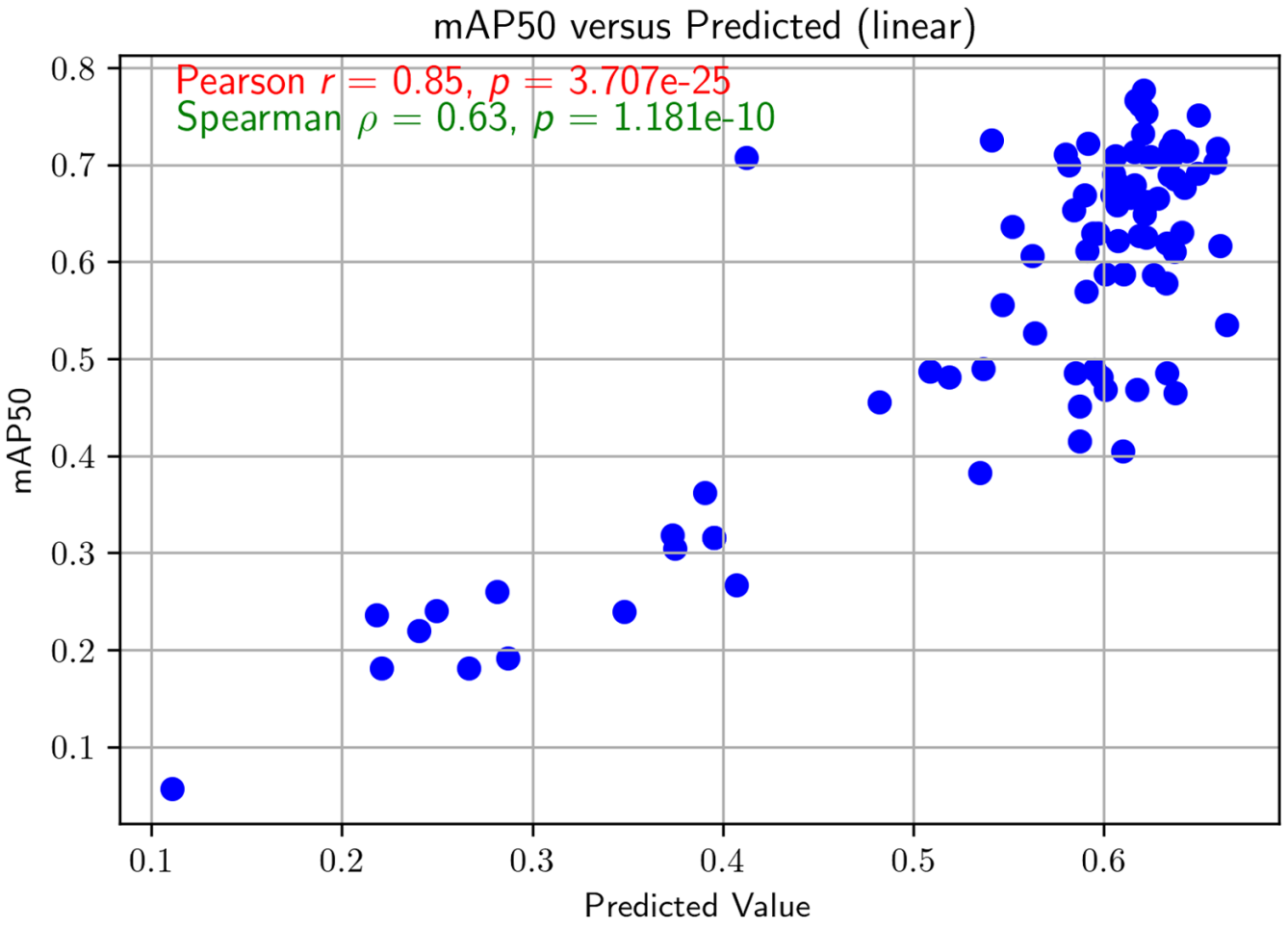

**Fig. 5** Validation datapoints with SDQM values calculated from **linear** regression coefficients versus YOLO11n mAP50 scores.

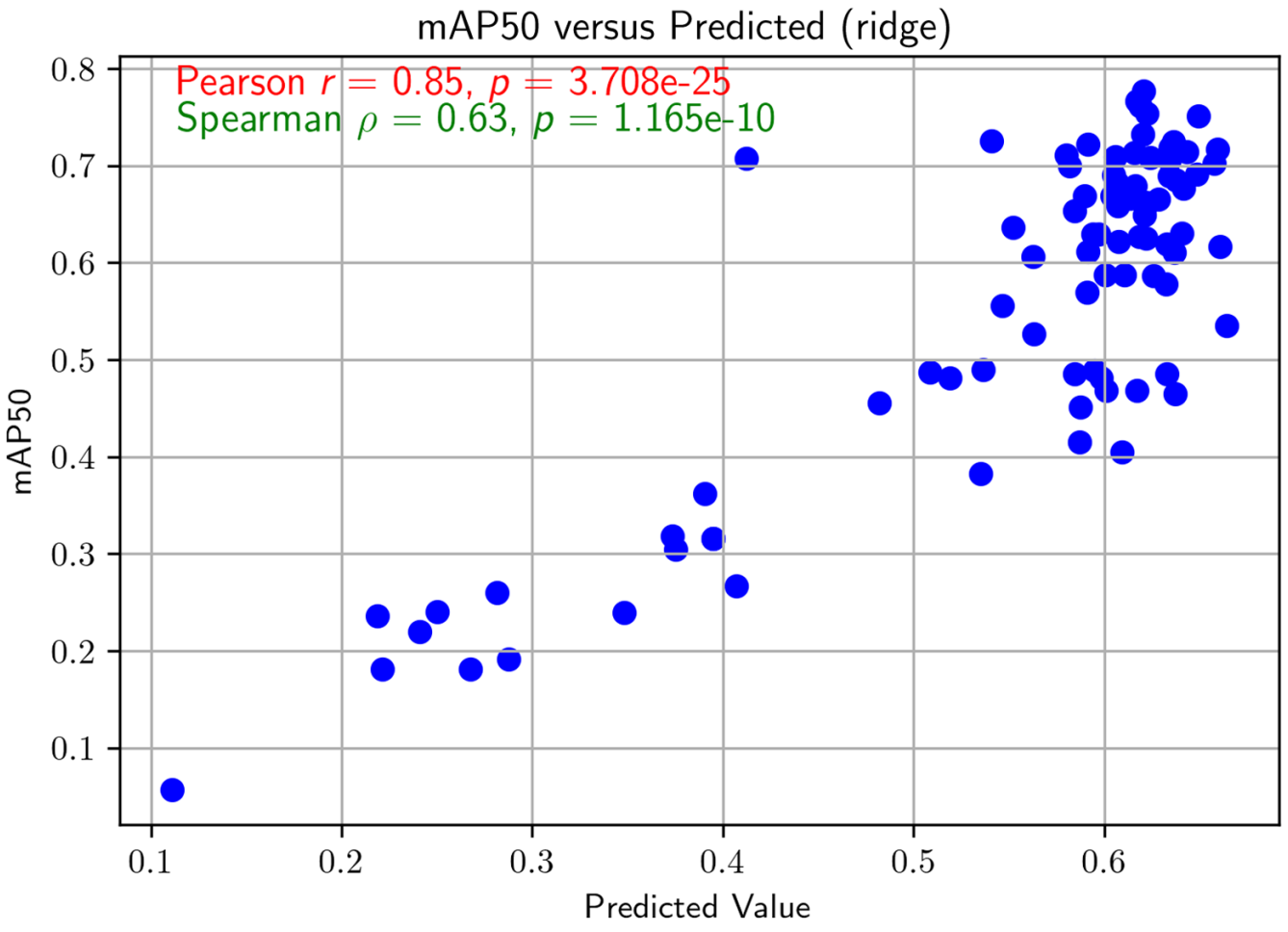

**Fig. 6** Validation datapoints with SDQM values calculated from **ridge** regression coefficients versus YOLO11n mAP50 scores.

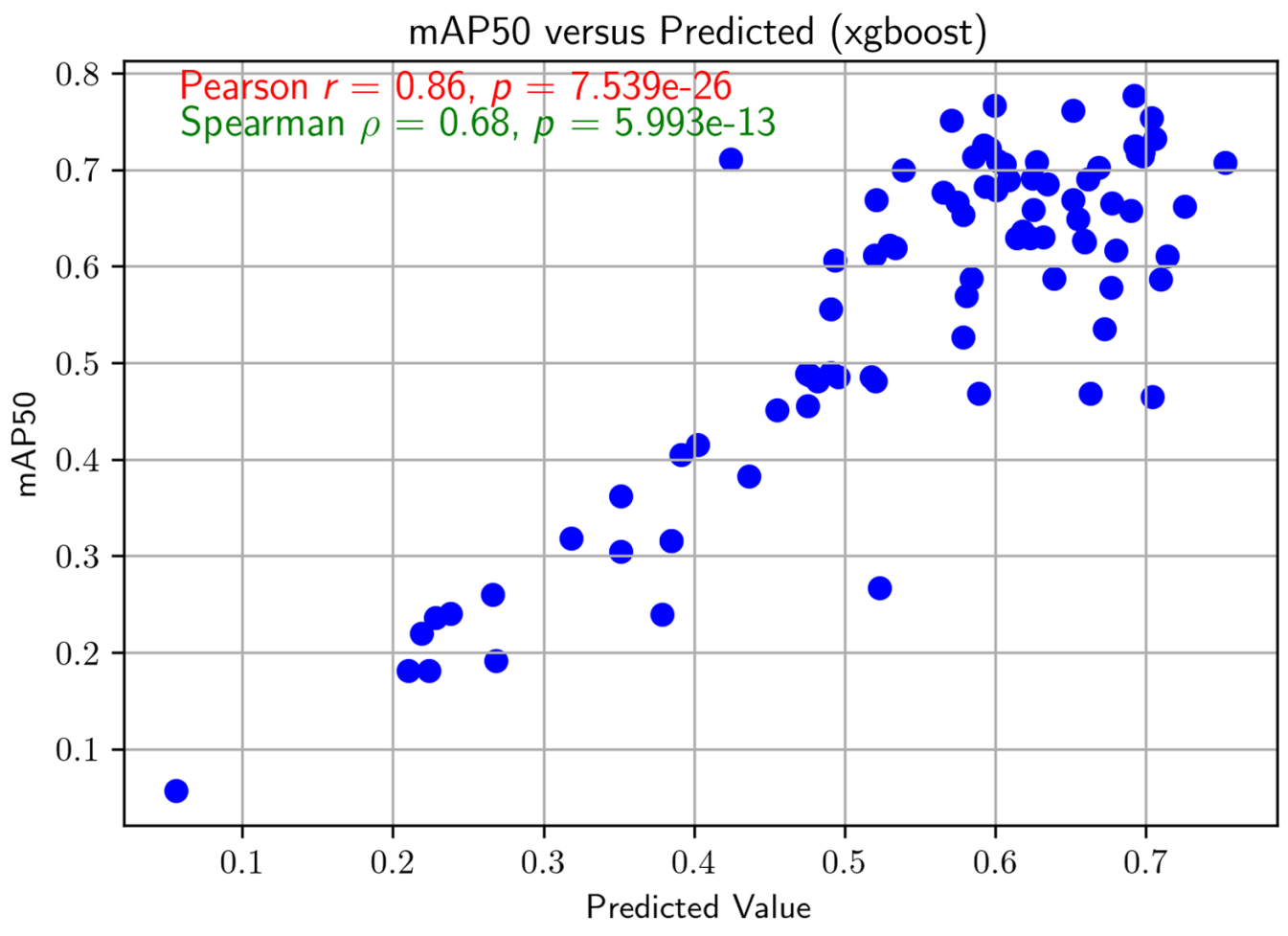

**Fig. 7** Validation datapoints with SDQM values calculated from **XGBoost** coefficients versus YOLO11n mAP50 scores.

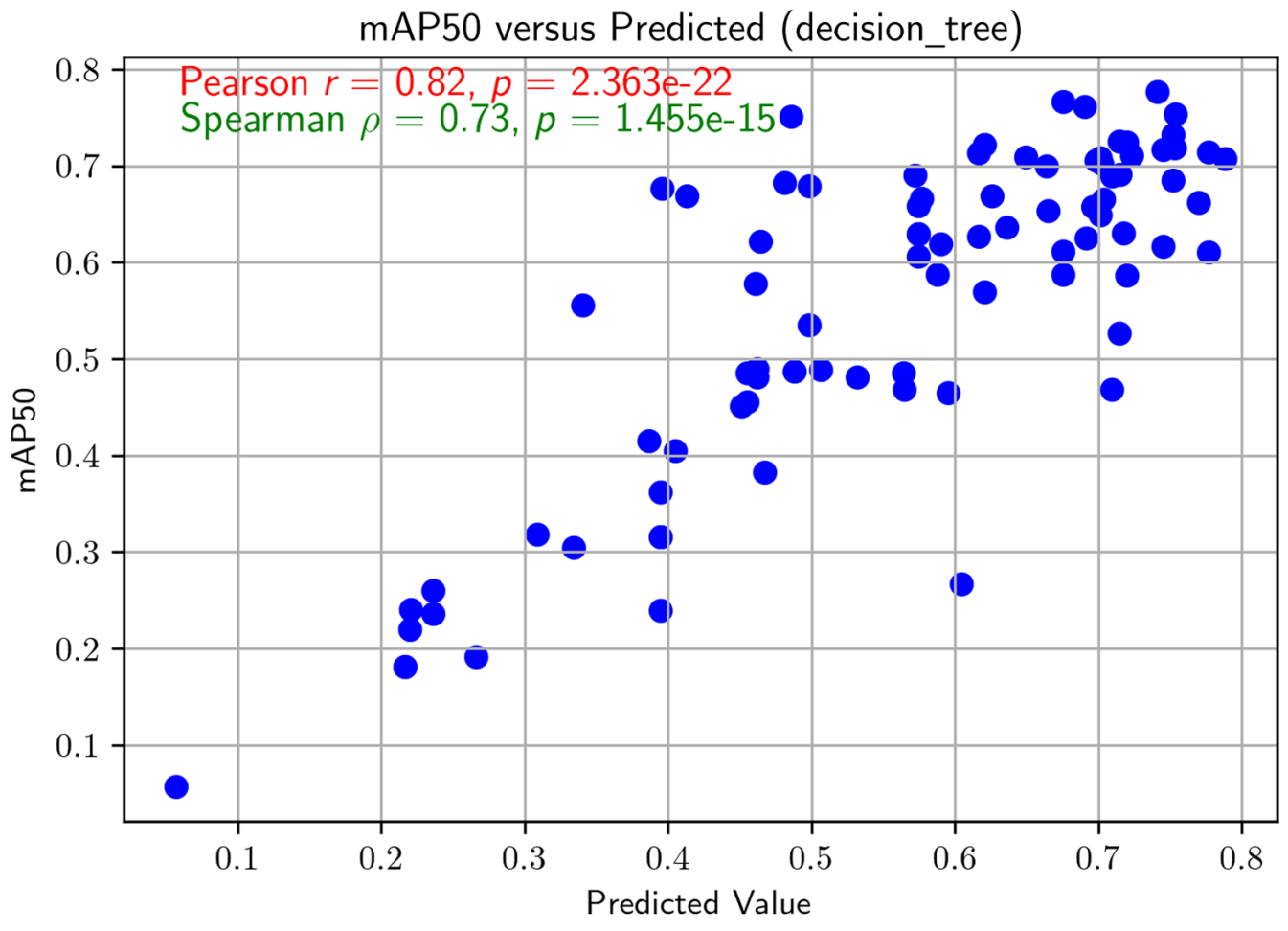

**Fig. 8** Validation datapoints with SDQM values calculated from **decision tree** coefficients versus YOLO11n mAP50 scores.

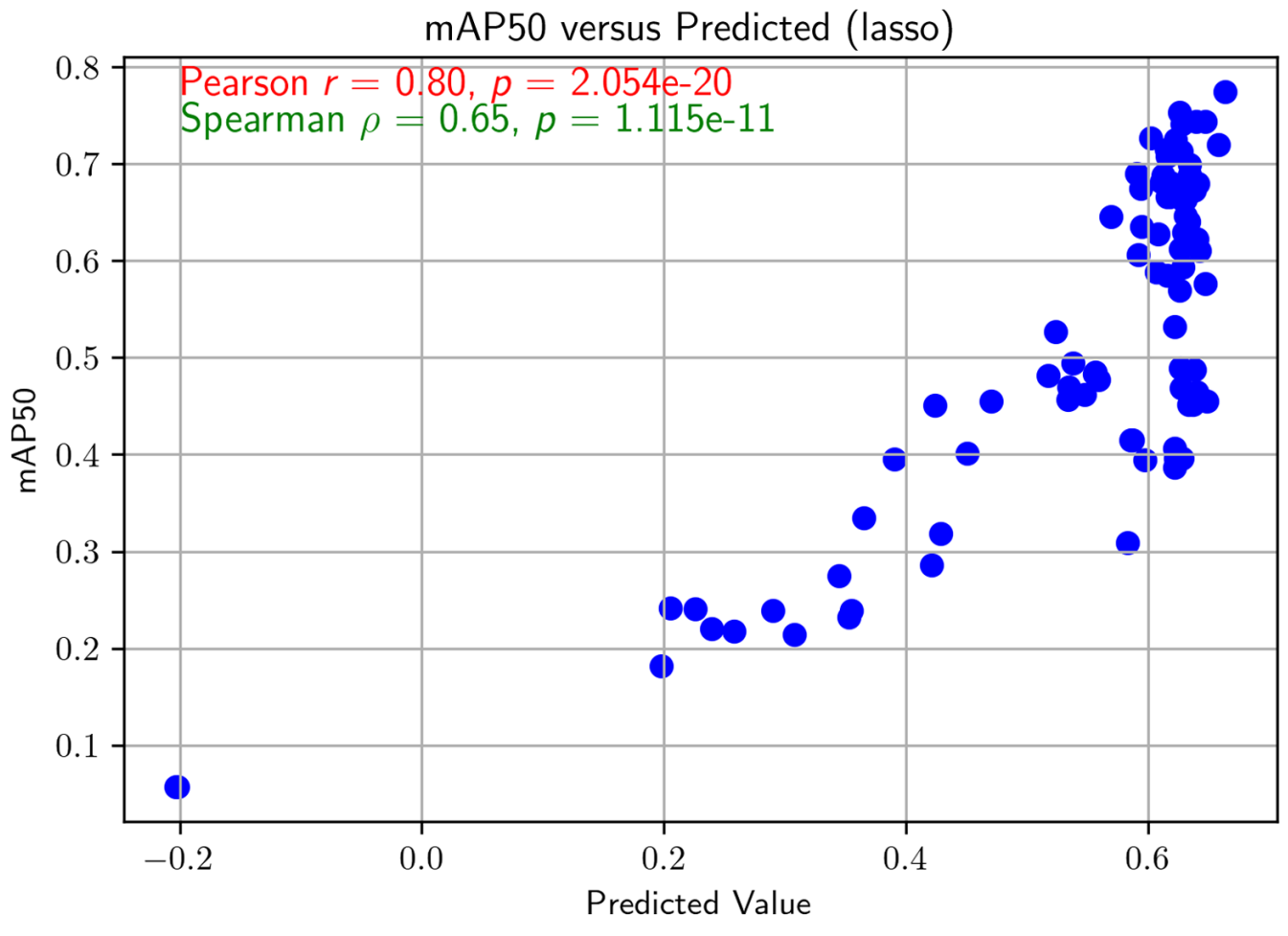

**Fig. 9** Validation datapoints with SDQM values calculated from **lasso** regression coefficients versus YOLO11n mAP50 scores.

Decision tree regression, displayed in Fig. 8, achieves adequate performance on our data. The added robustness of Random Forest achieved by introducing randomness may explain the deficit in the performance of decision tree regression.

Lasso regression, although showing promise over linear regression with an added regularization term, failed to meet our performance needs as shown in Fig. 9. Lasso applies $L1$ regularization to enforce sparsity (setting coefficients to zero). Because the submetrics are dense (i.e., every submetric selected contributes a small amount to the prediction), Lasso aggressively removes necessary information by zeroing out features, leading to underfitting.

## 9 Appendix D: Submetric Regression Analysis

We perform quadratic analysis to combine submetrics into subcomponents, removing terms with $P$ values $> 0.05$ for statistical significance. This reduces the number of terms needed.

For Bounding Box Match, after determining energy distance to be the most effective statistical method for comparison based on $P$ values, we analyze the remaining subcomponents. Removing subcomponents from the regression used for combination with a $P$ value $> 0.05$, we are left with the energy distance between aspect ratio, size, and area distributions. Also left included are the square of the aspect ratio and size comparisons, as well as the interaction term between aspect ratio and size comparisons.

Applying the same process to the regression of the subcomponents of Pixel Intensity Match after A-D Statistic was selected, the green pixel-value subcomponent and the square of the red pixel-value subcomponent along with the interaction term between them are left. The blue pixel-value terms were completely removed, suggesting the blue-pixel value match is not as important for predicting mAP.

The last of the submetrics utilizing nonparametric statistical methods, Label Overlap, contains only one subcomponent. Our regression analysis determined that the square of the K-S Statistic value was the most useful variable for this submetric.

MAUVE, which was dropped from the final metric, consists of MAUVE and MAUVE*—the latter being a smoothed version. We use $(\text{MAUVE}^*)^2 + (\text{MAUVE})(\text{MAUVE}^*)$ to combine these subcomponents. We found that Frontier Integral, which correlated highly with MAUVE, was most effective with only its smoothed variant as a single linear term.

Clusterability (removed from the final SDQM metric), consisting of subcomponents with ($l$) and without ($c$) a logarithm applied, reduced to the following regression equation: $c^2 + l^2 + cl + c + l$.

$\mathcal{V}$-information, the submetric with the highest correlation with mAP50, reduced to predictive confidence, conditional fusion, and the interaction term between IoU $\mathcal{V}$-information and conditional fusion.

Submetrics with only one subcomponent, denoted $x$, include spatial distribution difference, $\alpha$-precision, $\beta$-recall, and authenticity (authenticity was removed from the final metric). All besides $\alpha$-precision, which uses a single linear term, use a linear term and a quadratic term $(x + x^2)$.

---

### Disclosures

We acknowledge that there are no conflicts to declare.

### Code and Data Availability

The code used in our study is publicly available at https://github.com/ayushzenith/SDQM. The WASABI dataset can be requested from the corresponding author. The DIMO dataset is accessible at https://pderoovere.github.io/dimo, and RarePlanes can be found at https://www.iqt.org/library/the-rareplanes-dataset.

### Acknowledgments

We acknowledge the AFRL Internship Program to support Ayush Zenith, Arnold Zumbrun, and Neel Raut's work. This material is based upon work supported by the Air Force Research Laboratory (Grant No. FA8750-20-3-1004). Any opinions, findings, conclusions, or recommendations expressed in this publication are those of the authors and do not necessarily reflect the views of the U.S. Air Force. The U.S. Government is authorized to reproduce and distribute reprints for Governmental purposes notwithstanding any copyright notation thereon. Approved for Public Release; Distribution Unlimited: AFR/PA Case No. AFRL-2024-6945.

**Ayush Zenith** is a researcher in computer vision and robotics. He is currently a perception researcher at Tera AI. He is an alumnus of the Generalizable Robotics and Artificial Intelligence Laboratory (GRAIL) at Northeastern University and the Air Force Research Laboratory (AFRL) in Rome, NY. His research interests include synthetic data generation, geospatial intelligence, world models, 3D perception, and high-level robotic planning.

**Arnold Zumbrun** is an undergraduate computer science student at Binghamton University, specializing in artificial intelligence (AI). He conducts AI research at the Air Force Research Laboratory (AFRL) in Rome, NY. His research focuses on understanding and metricizing the quality of synthetic data for use in object detection.

**Neel Raut** is an undergraduate computer science student at Northeastern University, specializing in artificial intelligence and machine learning. He is an alumnus of the Air Force Research Laboratory (AFRL) in Rome, NY. His research interests include object detection and segmentation, synthetic data generation, reinforcement learning, and quantitative applications of machine learning.

**Jing Lin** received her PhD in mathematics from the University of South Florida in 2022. She is currently a research mathematician at the Air Force Research Laboratory (AFRL) in Rome, NY. Her research interests include machine learning, synthetic data generation, data quality metrics, adversarial attacks, and data poisoning attacks.